# "Mirror" Large Language Model Evaluations of Depression are Criterion Contaminated

Tong Li, Rasiq Hussain, Mehak Gupta, and Joshua R. Oltmanns

Author's note:

Tong Li, M.S., Division of Computational & Data Sciences (DCDS), Washington University in St. Louis; Rasiq Hussain, M. S., and Mehak Gupta, Ph.D., Department of Computer Science, SMU; Joshua R. Oltmanns, Ph.D., Department of Psychological & Brain Sciences, Washington University in St. Louis.

The authors would like to thank Jocelyn Brickman, Cameron Mills, Anton Mays, Megan Mathur, and Pooja Heragu for their dedicated efforts in collecting the data for this project.

Correspondence should be addressed to Tong Li, Division of Computational & Data Sciences (DCDS), Washington University in St. Louis, St. Louis, MO. Email: li.tong@wustl.edu.

Abstract

Large Language Model (LLM) studies that use language responses elicited from depression assessments to predict scores on those same assessments often report near-perfect prediction of depression. We refer to these as "Mirror" evaluations and demonstrate an applied case of criterion contamination. $N = 110$ participants completed *both* structured diagnostic depression interviews (Mirror condition) and life history interviews ("Non-Mirror" condition). LLMs were prompted to predict depression scores in each condition. As expected, Mirror evaluations were near-perfect. However, Non-Mirror evaluations also displayed prediction sizes considered *outstanding* in psychology. Further, both Mirror and Non-Mirror predictions correlated with Patient Health Questionnaire-9 scores at similar sizes ($r \approx .54$), suggesting the Mirror condition's advantage collapses when predicting an independent depression measurement. Topic modeling revealed differing depression-related themes across interview types. Mirror evaluations are better considered as *reliability* evaluations than as *validity* evaluations. Incorporating Non-Mirror approaches in LLM depression assessment may support more valid and clinically-relevant applications.

*Keywords:* large language models, psychological assessment, psychopathology, depression, reliability, validity, criterion contamination

# “Mirror” Large Language Model Evaluations of Depression are Criterion Contaminated

Large language models (LLMs) have the potential to increase depression assessment validity, providing a scalable assessment tool easily implemented into routine clinical settings (Stade et al., 2024). LLM-based assessment of depression is accurate using clinical language samples (Omar & Levkovich, 2025), with some studies demonstrating very high prediction accuracies (Kebe et al., 2025). However, in psychological assessment, high predictive accuracy does not necessarily equal validity. A central validity question is whether the resulting prediction generalizes beyond the elicitation context (Alaa et al., 2025; Brickman et al., 2025; Chancellor & De Choudhury, 2020). We propose that language-based AI models for depression assessment studies should distinguish between Mirror and Non-Mirror evaluations. This distinction is not intended as a new machine-learning theory of generalization, but as a psychometric distinction about the relation between the language sample and the criterion being predicted.

We define a Mirror evaluation as an LLM-based assessment setting in which the input language is elicited by questions that directly correspond to the same symptom criteria used to create the predicted score. In contrast, in a Non-Mirror evaluation, the same score is predicted from language not explicitly structured around those criteria, such as life-history narratives or other routine clinical language. The term “Mirror” is therefore used descriptively: the predictor language mirrors the measurement procedure that generates the criterion (Figure 1a). Several recent studies illustrate this Mirror setting. In one study, the Montgomery-Asberg Depression Rating Scale (MADRS) (Montgomery & Asberg, 1979) was administered to $N = 236$ patients (Kebe et al., 2025). LLaMA and Qwen 2.5 were then prompted to provide predictions on the MADRS—using language responses to the MADRS that mirrored the MADRS questions (Kebe et al., 2025). $R^2$ for the full interview ratings reached .69. In another study, the Patient Health

Questionnaire-9 (PHQ-9) (Thombs et al., 2014) was administered to *N* = 393 community adults, and a proprietary algorithm predicted PHQ-9 scores *from language responses to PHQ-9 questions*, yielding a Pearson correlation of *r* = .73 between predicted and actual scores and over 90% agreement with either self-report or clinical opinion (Weisenburger et al., 2024). A third study administered the PHQ-9 to *N* = 963 patients online (Ganesan et al., 2024). Patients were then asked to provide detailed writing specifically about their depression over the last two weeks. GPT-4 Turbo was prompted to predict PHQ-9 scores from the writings about depression. While this third study did not specifically use language responses to the PHQ-9, language responses to queries specifically asking to detail depression experiences over the same time frame were used. GPT-4 Turbo predictions correlated *r* = .70 with actual PHQ-9 responses. Across these examples, the common thread was that the input language was elicited within a framework explicitly designed to match the assessment criteria (that were then predicted).

Mirror evaluation studies show very large effect sizes that are unprecedented in cross-method convergence studies in psychology (Meyer et al., 2001), and are larger than many *mono*method convergent validity estimates in psychology (e.g., correlations of one self-report with a different self-report of the same construct). The effects seen in Mirror studies approach perfect prediction before even considering correction for attenuation due to imperfect reliability, which would increase the effect size further. These large effects are understandable from a machine-learning perspective because the predictor language is closely aligned with the assessment target. However, this same alignment raises a validity concern in psychological assessment. Psychological measures contain measurement error, and when model predictions approach the reliability limits of the criterion itself, high performance may reflect overlap between the predictor and criterion rather than generalizable depression detection. This concern is an applied instance of “criterion contamination” (Bellows, 1941; Goldberg, 1965), which occurs when the outcome that a model is designed to predict is partly derived from the same information used as the predictor. Criterion contamination artificially inflates the effect sizes and misrepresents true predictive power. “When criterion diagnoses are derived from the same source of information that generates one of the predictor variables, then that predictor and the criterion are confounded, and this should produce artificially high estimates of the predictor’s validity or accuracy” (Hunsley & Meyer, 2003, p. 452). It is therefore recommended that statistical models are not developed to predict the data they were developed on: “To avoid criterion contamination, when deriving statistical-prediction rules that use psychological test information and history, interview, and observation data, one cannot base criterion scores on the same history, interview, and observation data that was given to the statistical-prediction rule” (Garb, 2000, p. 36). Due to potentially inflated results in Mirror studies from criterion contamination, findings from Mirror evaluations may not generalize across different types of language samples. That is, they may generalize primarily to language samples that reproduce the structure and content of the original depression assessment.

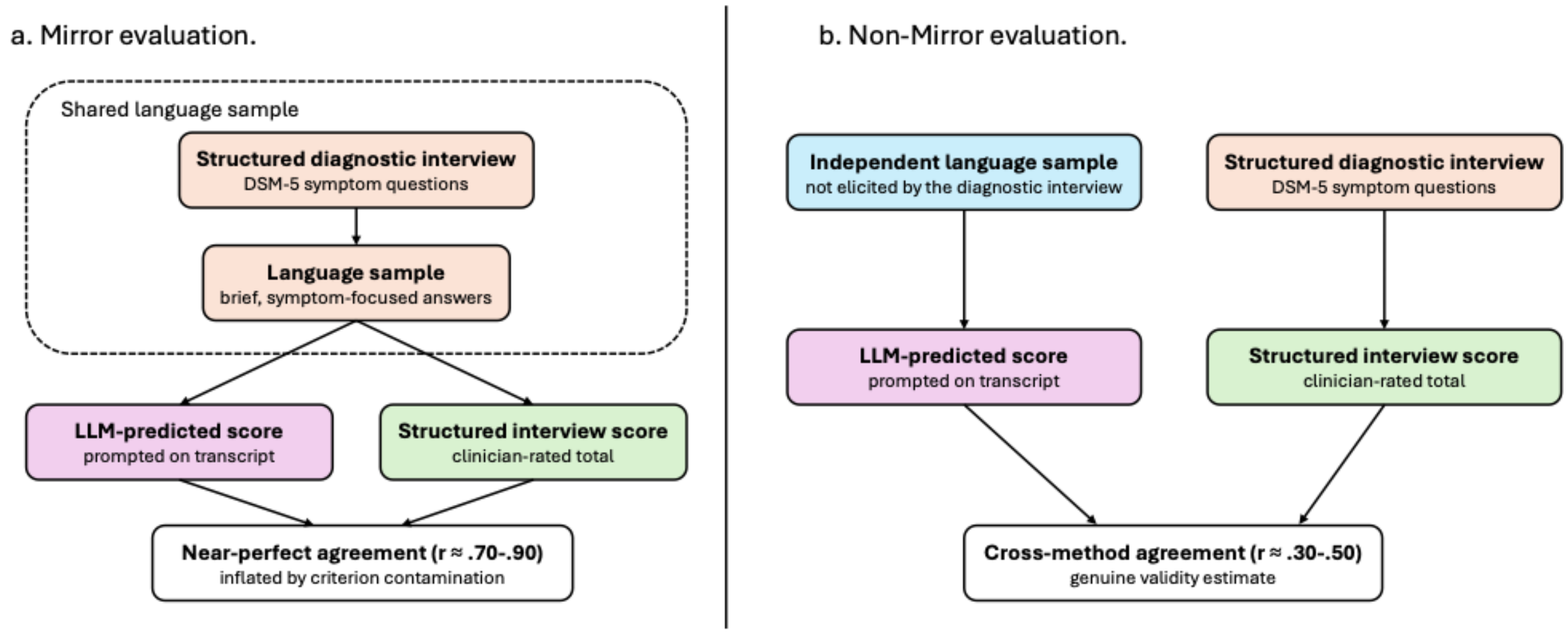

**Figure 1 Mirror versus Non-Mirror evaluation of language-based AI depression assessment.** (a) In a Mirror evaluation, a single language sample — elicited by the structured

diagnostic interview — serves as the source of both the LLM-predicted score and the clinician-rated criterion score. The dashed enclosure denotes this shared language sample: because predictor and criterion derive from the same body of interview language, their agreement is inflated by criterion contamination. (b) In a Non-Mirror evaluation, the LLM prediction derives from a separate language sample (not elicited by the diagnostic interview), while the criterion score derives from the structured diagnostic interview. Because the predictor and criterion have independent origins, no shared language source exists, and the observed agreement constitutes a genuine cross-method validity estimate. Correlation magnitudes reflect results typical of prior Mirror evaluations and of cross-method agreement between distinct assessment methods in psychological research, respectively. In the present study, the Non-Mirror language sample was a life history interview not structured around DSM-5 criteria.

Importantly, there are also studies demonstrating depression can be predicted from language samples that do *not* mirror structured depression assessments. These Non-Mirror approaches (Figure 1b) include those developed from unstructured or semi-structured clinical interviews, case histories, life narrative interviews, social media language, and other language samples not explicitly elicited by the target assessment. For example, one study used machine learning to predict responses to the IPIP Depression scale using a collection of Facebook status updates from $N = 348$ users (Park et al., 2015). This was a Non-Mirror evaluation because the language sample (Facebook status updates) was not language responses to the Depression scale, nor were the users instructed to describe their depression. The Pearson $r$ correlation between the model's Depression predictions and responses to the Depression scale was $r = .37$. However, while this study is promising, the models may have limited clinical utility: social media language samples are not spoken language and are not able to be collected from all clinical patients.

In another study, the RoBERTa language model (Liu et al., 2019) was fine-tuned to predict Neuroticism from spoken language in life narrative interviews with $N = 1{,}409$ community older adults (Oltmanns et al., 2026). In the interviews, participants were asked to divide their life stories into 3-4 chapters and were subsequently asked about best and worst characters in their lives, high and low points, and a turning point. The NEO-Personality Inventory-Revised (NEO-PI-R) (Costa & McCrae, 1992) Neuroticism scale (which contains a Depression subscale) was administered and used as a label for training. Importantly, these were not language responses to the Neuroticism scale, nor were participants asked about depression symptoms. The Pearson *r* correlation between the model Neuroticism predictions and the responses to the Neuroticism scale was $r = .43$. While the effect sizes of the prediction models were smaller than in Mirror evaluations, they are still in a range that is considered large in psychology (Funder & Ozer, 2019) (e.g., $r \approx .40$) and indicate that assessment scores can be predicted from language not directly elicited by the assessment itself. The implications are that *various* language samples may be used to predict depressive symptoms, which may lead to more generalizable, scalable, and clinically relevant language models of depression in the future.

Taken together, these literatures suggest the two evaluation types measure different things. A Mirror evaluation provides a *reliability* test rather than a validity test. How well can the LLM, as a second rater, agree with the primary clinician rating? In this sense, Mirror evaluations provide insight into the potential of using LLMs as secondary raters or reliability checks. Because agreement in the Mirror condition reflects overlap between the predictor and the outcome, it provides little new information beyond the diagnostic criteria themselves. As reliability tests, they approach the reliability of the criterion itself (e.g., *r*'s ≈ .70 to .90).

A Non-Mirror evaluation, in contrast, is a true validity test. How well can the LLM identify signs of depression from a separate language source, for example routine clinical language or life history? Non-Mirror coefficients should not be judged against Mirror (reliability) coefficients because that would compare a validity estimate to a reliability estimate. Non-Mirror tests such as those described above are truly *cross*-method (e.g., LLM ratings of a life history interview versus clinician ratings of a structured diagnostic interview). Monomethod convergent validity coefficients (e.g., the correlation between two self-report scales of depression) are typically $r > .50$, and cross-method validity coefficients $r > .30$ (Clark & Watson, 2019; Meyer et al., 2001). By these standards, findings from Non-Mirror evaluations (*r*'s .30-.40; Oltmanns et al., 2026; Park et al., 2015) already cross the threshold for strong cross-method convergent validity test results. Non-Mirror contexts are also where LLMs can contribute something new: LLMs may identify novel linguistic and contextual correlates of depression when evaluating language from life history interview sections (e.g., current functioning, social and occupational history). However, direct evidence on the inflation of Mirror evaluations is lacking. Existing Mirror and Non-Mirror evaluations come from different samples, criterion measures, language modalities, and modeling approaches. Isolating the contribution of criterion contamination requires a within-sample comparison in which the participants, the criterion, and the models are held constant while only the language sample varies.

To address this question, the present study directly compares Mirror and Non-Mirror LLM evaluations of depression within the same sample. We analyzed $N = 110$ transcripts from participants who each completed two distinct interviews: a structured diagnostic interview designed around DSM-5 criteria for major depressive disorder, and a life history interview resembling a routine clinical intake. Transcripts were automatically generated, anonymized, and segmented, providing parallel inputs for both evaluation approaches. Large language models (GPT-4 Turbo, GPT-4o, and LLaMA-3 70B) were prompted to infer depressive symptoms from each transcript type, using a reasoning framework that emphasized evidence grounding, confidence estimation, and structured output. We then evaluated performance with standard classification and regression metrics, and further probed the semantic structure of model-selected utterances using topic modeling to understand when predictions align or diverge from clinical ground truth.

The primary contribution of the study is to provide an applied empirical demonstration of criterion contamination in language-based depression assessment, while also evaluating whether depressive symptoms can be inferred from less structured clinical language. Prior studies have been limited by reliance on social media text and use of labels with problematic construct validity (Chancellor & De Choudhury, 2020). Strengths of the present study include a rich, spoken language sample of *two distinct types of clinical interviews* (structured diagnostic interview and semi-structured life history) to be directly compared in the same sample. Further, the present study includes a construct-valid, structured interview-based depression assessment, which is needed in language AI model development to improve construct validity of model predicted scores (Chancellor & De Choudhury, 2020). We expected Mirror evaluations to approach the reliability ceiling of the criterion, and Non-Mirror evaluations to yield smaller coefficients in the cross-method validity estimate range.

# Method

## Participants

Participants were recruited from a university setting through the psychology department subject pool and flyers posted on campus. The final analytic sample included 110 participants who completed the study procedures and had usable audio recordings and transcripts for both interview conditions. Demographic, academic, socioeconomic, and depression-symptom characteristics of the sample are reported in Appendix Table S1.

## Procedure

Participants provided informed consent and completed a questionnaire battery on a computer in the lab. Life history and structured diagnostic interviews were then conducted as the participants were seated at a table across from an interviewer. Interviewers were clinical psychology PsyD students trained in interviewing by the senior author and as part of their graduate training program. Audio was captured using two supercardioid dynamic microphones, one for the interviewer and one for the participant, with each routed to a separate channel on a digital multitrack recorder. Directly following the life history interview, a structured interview module for major depressive disorder was conducted, in addition to modules for several other disorders. Participants were then debriefed and provided a list of mental health resources.

**Measures**

*Life History Interview.* Participants were administered a broad interview developed to resemble an initial appointment/intake/patient history interview, but it was adapted for college students. Development of this interview consisted of reviewing intake clinical interviews from several academic and clinical settings and online resources, and formulating an interview consisting of five sections: Section 1 asked five broad questions about education and employment history, Section 2 asked nine broad questions about current relationships and support systems, Section 3 asked five broad questions about sleep, eating, and exercise, Section 4 asked seven broad questions about physical and mental health history, and Section 5 asked five broad questions about strengths, coping, and the future.

*Structured Diagnostic Interview.* Ten structured clinical interview questions were adapted from the Mini-International Neuropsychiatric Interview DSM-5 diagnostic criteria for major depressive episode (Sheehan et al., 1998). Questions focused on the past two weeks, and asked about feeling sad empty and hopeless, much less interested in things, decreased or increased appetite, trouble sleeping, moving more slowly or becoming more fidgety, feeling tired, feeling worthless or guilty, having trouble concentrating or making decisions, repeatedly thinking about death, and whether these symptoms caused significant distress or problems in important areas of life. Although questions are “yes” or “no,” participants often added brief descriptions to their responses, which were included in the transcripts. Items were scored “0” not present or “1” present. Sum scores were created by totaling the ten items. Internal consistency was $\alpha = .66$. Interrater reliability for the 10 items ranged from ICC = .84 (being much less interested in most things or unable to enjoy things most of the time) to ICC = 1.0 (items related to talking or moving more slowly, or being more restless or fidgety; and feeling worthless or guilty almost every day), with a median of ICC = .93.

*Self-Report Questionnaire.* Participants completed the Patient Health Questionnaire-9 (Thombs et al., 2014) to assess subjective depressive symptoms. The measure assesses nine symptoms analogous to the DSM-5 major depressive disorder criteria over the last two weeks. Items are rated from 0 (*not at all*) to 3 (*nearly every day*). A sum score was computed. Ranges 1-4 indicate minimal depression, 5-9 mild, 10-14 moderate, 15-19 moderately sever, and 20-27 severe depressive symptoms. In the present study, PHQ-9 scores were used as an independent self-report measure of depressive symptoms for evaluating the external convergence of Mirror and Non-Mirror LLM conditions.

*External Validation Criteria.* To compare the validity across mirror and non-mirror predictions, we administered eight external criterion measures. The Generalized Anxiety Disorder scale (GAD-7; Spitzer et al., 2006) is a 7-item self-report measure of anxiety symptoms over the past two weeks, with each item rated from 0 ("not at all") to 3 ("nearly every day"). The PTSD Checklist for DSM-5 (PCL-5; Blevins et al., 2015) is a 20-item self-report measure of DSM-5 posttraumatic stress symptoms, with each item rated from 0 ("not at all") to 4 ("extremely"). The Brief Symptom Inventory (BSI; Derogatis & Melisaratos, 1983) is a self-report measure of psychological distress; we administered three of its subscales—Somatization, Paranoid Ideation, and Psychoticism—with items rated from 0 ("not at all") to 4 ("extremely"). The Drug Abuse Screening Test (DAST; Skinner, 1982) is a 10-item self-report screen for drug-use problems, with items answered in a yes/no format. The Alcohol Use Disorders Identification Test (AUDIT; Saunders et al., 1993) is a 10-item self-report screen for hazardous and harmful alcohol use, with most items rated from 0 to 4. The Satisfaction With Life Scale (SWLS; Diener et al., 1985) is a 5-item self-report measure of global life satisfaction, with items rated from 1 ("strongly disagree") to 7 ("strongly agree"). The Elemental Psychopathy Assessment Super-Short Form (EPA-SSF; Collison et al., 2016) is an 18-item self-report measure of psychopathic traits derived from the Five-Factor Model, with one item per EPA subscale and items rated from 1 ("strongly disagree") to 5 ("strongly agree"). The Borderline Pattern Scale (BPS; Oltmanns & Widiger, 2019) is a 12-item self-report measure of borderline personality pathology, with items rated 1 ("strongly disagree") to 5 ("strongly agree"). The SOURCE measure (Lawton, 2014) is a 4-item self-report measure of social dysfunction, with items rated from 1 ("strongly disagree") to 5 ("strongly agree").

**Data Preprocessing**

Audio recordings were transcribed using the Whisper automatic speech recognition (ASR) model, which supported speaker diarization to differentiate interviewer and interviewee utterances. Transcripts were initially stored in .srt format with timestamp and speaker metadata. Given the sensitive nature of clinical interviews, we implemented a two-stage anonymization pipeline. Named entity recognition (NER) was applied using the spaCy *en_core_web_trf* model to identify and redact personal identifiers, including names of people, organizations, and locations. In parallel, a supplementary list of custom sensitive terms (e.g., local location names) were applied, ensuring comprehensive de-identification while preserving conversational context.

Following de-identification, transcripts were algorithmically segmented into life history interviews and structured diagnostic interviews. The life history component came first in all interviews and is distinguished based on questioning in five areas (introduced in Measures). The structured interview came after the life history and was identified by the presence of characteristic phrases (e.g., "structured part of the interview", "mental health symptoms") and corroborated by patterns of short affirmative or negative responses. Because the structured interviews assessed several forms of mental disorder, further segmentation was performed to isolate the depression-focused structured interview component. All processing steps were fully automated across interview files to ensure standardization and scalability. Processed outputs included fully anonymized transcripts, as well as life history interviews and structured diagnostic interviews.

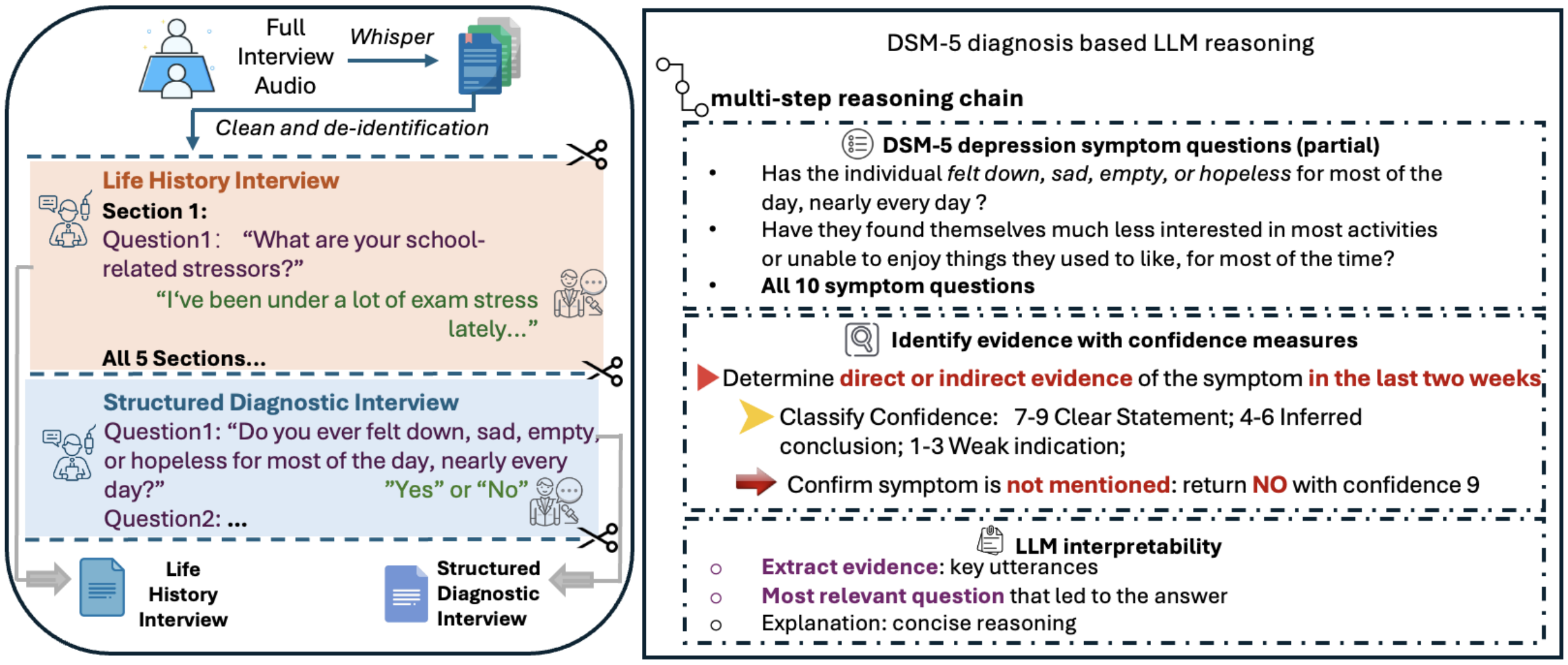

**Figure 2. Evidence-grounded LLM inference pipeline for Mirror vs. Non-Mirror condition depression diagnosis.** This Figure illustrates the overview of the data preprocessing and LLM reasoning pipeline. (Left) Audio-recorded clinical and structured interviews were transcribed

using Whisper and subsequently anonymized via named entity recognition and sensitive keyword filtering. Structured interview segments were further segmented to isolate content relevant to the five-section DSM-5 depression assessment, while excluding unrelated segments such as anxiety and trauma. All segmented interview data were subsequently reviewed and corrected to ensure accurate separation between structured diagnostic (Mirror) and life-history (Non-Mirror) interview content. (Right) A multi-step prompting framework guided LLM-based reasoning to infer the presence of DSM-5 depressive symptoms. For each question, the model identified direct or indirect evidence from the past two weeks, classified confidence levels (clear, inferred, or weak), and provided a structured output including a key utterance, relevant context, and concise explanation. This framework enabled transparent and evidence-grounded symptom inference.

**LLM reasoning**

Each anonymized transcript was analyzed using a structured large language model (LLM) reasoning framework designed to infer the presence or absence of depressive symptoms according to DSM-5 diagnostic criteria. The reasoning process was constructed to increase interpretability and reduce unsupported inferences.

Our approach employed a multi-step prompting strategy to guide the model through a sequential, evidence-based reasoning process (Figure 2). First, the model was instructed to identify direct or indirect evidence for each symptom, specifically restricted to the past two weeks. The model was then prompted to self-assess the strength of the evidence and assign a confidence score on a 1–9 scale. Clear, direct symptom reports received higher confidence scores (7–9), while inferential or ambiguous indications resulted in moderate scores (4–6). In the absence of supporting evidence, the model was explicitly instructed to predict "No" with the highest confidence score (9), discouraging speculative or unfounded conclusions. To further enhance transparency, the model extracted specific key utterances that supported its prediction and identified the most relevant interviewer question that prompted the discussion of the symptom. This structure enabled human reviewers to audit model outputs and verify that predictions were traceable to face-valid portions of the transcript.

The prompting framework was designed as a structured evidence-grounded procedure rather than an unconstrained diagnostic judgment. It required the model to first identify transcript evidence, then generate a symptom prediction, confidence score, supporting utterance, and brief explanation. (the full prompt and algorithm are illustrated in Supplemental Figure S1, Appendix). Prior research has demonstrated that grounding model outputs in explicit evidence substantially reduces hallucination rates, while prompting models to self-evaluate their confidence enhances the trustworthiness of model-generated clinical judgments (Dziri et al., 2022; Tao et al., 2024). Moreover, explicit extraction of supporting evidence has been emphasized as an essential component for building explainable and clinically actionable AI systems (Rajpurkar et al., 2022).

Inference was performed using three off-the-shelf LLMs: GPT-4 Turbo, GPT-4o, and LLaMA-3 70B. GPT-4 Turbo and GPT-4o were selected as widely used proprietary models with strong general language understanding and reasoning capabilities, whereas LLaMA-3 70B was included as a large open-weight comparison model. To support reliable processing across transcripts, we include automatic retry mechanisms for invalid outputs and API key rotation in the inference pipeline. We set temperature=0 to maximize reproducibility and max_tokens=2048 to ensure sufficient response length and avoid truncation. Other generation controls remain at default settings: top_p=1, frequency_penalty=0, presence_penalty=0, n=1, stream=false, stop=null, with seed, logit_bias, and user left unset. In short, we use deterministic decoding with a sufficiently large output budget while keeping the rest of the configuration unchanged. All model outputs were structured in JSON format, facilitating standardized downstream analysis and human review.

We adopted a dual-perspective evaluation to assess LLM performance in extracting depressive symptoms from life history and structured diagnostic interviews. This framework includes both classification metrics for symptom-level prediction accuracy, and regression metrics for overall symptom estimation (subject-level). Specifically, we report several standard classification metrics, including *Accuracy, Precision, Recall, F1-score, and Jaccard Similarity*, to evaluate the accuracy of LLM-generated binary predictions for each of the ten DSM-5 depression criteria. The *Jaccard Similarity* assesses the overlap between the predicted and true symptom sets, emphasizing intersection-over-union. It offers a more intuitive measure of overlap to capture set-level alignment between predicted and true symptom clusters. These metrics are computed both at the individual interview level and aggregated across all subjects for both Mirror and Non-Mirror conditions.

Beyond classification performance, we also aimed to capture how well the models reflect overall depressive symptom burden at the subject level. To this end, we computed the total number of predicted symptoms per subject and evaluated model estimates using regression-based metrics: *Mean Squared Error (MSE), Mean Absolute Error (MAE),* the *coefficient of determination* ($R^2$), computed as 1 minus the ratio of residual to total variance, *Pearson score* and *Spearman score.* While *MSE* and *MAE* quantify prediction accuracy from an error-minimization perspective, $R^2$ captures the proportion of variance in true symptom counts explained by model estimates, offering insight into the model's overall explanatory strength. The Pearson correlation measures linear relationships between two continuous variables—in this case, predicted vs. actual symptom totals. Alternatively, Spearman correlation, which assesses monotonic relationships based on ranked values when the underlying relationship is non-linear, provides a robust assessment.

**Topic Modeling**

We used topic modeling as an exploratory analysis to characterize the kinds of language LLMs relied on when making positive depression-symptom predictions. This analysis was intended to clarify how Mirror and Non-Mirror condition predictions differed semantically and to identify language patterns that contributed to correct versus incorrect Non-Mirror predictions. Specifically, we analyzed utterances that GPT-4 Turbo selected from the transcripts as supporting evidence for "yes" predictions on individual DSM-5 depression criteria. We first compared topic structures across interview formats, contrasting structured diagnostic interviews in the Mirror condition with life history interviews in the Non-Mirror condition. We then compared true-positive and false-positive utterances within the Non-Mirror condition to identify themes associated with accurate and inaccurate positive predictions; this comparison was restricted to the Non-Mirror condition because models in the Mirror condition produced too few false positives to support such an analysis.

Each utterance was preprocessed using standard NLP techniques, including contraction expansion, lowercasing, punctuation removal, and sentence segmentation. Utterances were encoded into high-dimensional semantic vectors using the all-MiniLM-L6-v2 sentence embedding model. To capture the local semantic structure while retaining sufficient representational capacity, we used UMAP (n_neighbors=10, n_components=5) for dimensional reduction. Semantic clustering was subsequently performed using HDBSCAN with a minimum cluster size of 55 to ensure thematic consistency. To enhance the interpretability of extracted topics, we used BERTopic for topic modeling, with a custom stopword list, which contains some additional filler and discourse words (e.g., uh, kinda, like, hmmm) frequently observed in spoken clinical interviews, and controlled vocabulary (e.g., exclude words that appear in more than 99% of the transcripts) to improve the quality of extracted topic keywords.

## Results

### Inflated Performance of Models in the Mirror Condition

Mirror evaluations demonstrated the expected high levels of predictive performance of DSM-5 criteria when applied to structured diagnostic interviews that directly assessed DSM-5 depression criteria. As shown in Table 1a, GPT-4 Turbo achieved an F1-score of 0.94, Jaccard similarity of 0.86 and $R^2$ of 0.80, with accuracy and correlation values consistently high (Acc. = 0.97; Pearson r = 0.90). GPT-4o and LLaMA-3 70B showed similar patterns, with F1-scores of 0.92 and 0.86 and $R^2$ values of 0.77 and 0.70, respectively. These results confirm that LLMs perform very well when the input language is elicited by the same diagnostic framework used to define the prediction target.

Table 1a. Model Performance in the Mirror Condition

| | | Acc. | Pre. | Rec. | F1 | Jac. | MSE | MAE | $R^2$ | $R^2$ CI (Lower) | $R^2$ CI (Upper) | Pearson $r$ | $r$ CI (Lower) | $r$ CI (Upper) | Spearman $r$ | $r$ CI (Lower) | $r$ CI (Upper) |
|---|---|---|---|---|---|---|---|---|---|---|---|---|---|---|---|---|---|
| Mirror | GPT-4 | **0.97** | **0.90** | 0.88 | **0.94** | **0.86** | **0.67** | **0.32** | **0.80** | **0.68** | **0.87** | **0.90** | **0.80** | **0.93** | **0.91** | 0.78 | **0.93** |
| | GPT-4o | 0.96 | 0.86 | 0.86 | 0.92 | 0.83 | 0.76 | 0.38 | 0.77 | 0.61 | 0.86 | **0.90** | 0.78 | 0.92 | 0.90 | 0.77 | 0.91 |
| | LLaMA3-70B | 0.93 | 0.66 | **0.92** | 0.86 | 0.70 | 1.01 | 0.62 | 0.70 | 0.59 | 0.82 | 0.88 | 0.77 | 0.90 | 0.87 | **0.79** | 0.92 |

Table 1b. Model Performance in the Non-Mirror Condition

| | | Acc. | Pre. | Rec. | F1 | Jac. | MSE | MAE | $R^2$ | $R^2$ CI (Lower) | $R^2$ CI (Upper) | Pearson $r$ | $r$ CI (Lower) | $r$ CI (Upper) | Spearman $r$ | $r$ CI (Lower) | $r$ CI (Upper) |
|---|---|---|---|---|---|---|---|---|---|---|---|---|---|---|---|---|---|
| Non-Mirror | GPT-4 | **0.79** | **0.37** | 0.40 | 0.52 | 0.34 | **2.48** | 1.23 | **0.27** | **0.10** | **0.44** | 0.57 | 0.46 | 0.69 | 0.56 | 0.44 | 0.64 |
| | GPT-4o | 0.76 | 0.33 | **0.44** | 0.53 | 0.33 | 2.79 | **1.17** | 0.16 | 0.08 | 0.38 | 0.56 | **0.49** | 0.69 | 0.57 | 0.39 | 0.67 |
| | LLaMA3-70B | 0.78 | 0.35 | 0.40 | **0.55** | **0.37** | 2.81 | 1.23 | 0.16 | 0.09 | 0.42 | **0.63** | 0.47 | **0.76** | **0.59** | 0.42 | 0.69 |

**Table 1.** Model performance comparison. Performance metrics across structured diagnostic (Mirror condition) and life history (Non-Mirror condition) interviews for GPT-4 Turbo, GPT-4o

and LLaMA models. Each column reports standard classification and regression metrics for binary symptom prediction (per DSM-5) under different interview conditions. Bold and italicized underlined words are the best predictions of the different models for Mirror and Non-Mirrored conditions, respectively. These metrics collectively evaluate model performance in both classification and regression tasks, with Accuracy (Acc.), Precision (Pre.), Recall (Rec.), F1 score (F1), and Jaccard similarity (Jac.) reflecting classification quality, while Mean Squared Error (MSE), Mean Absolute Error (MAE), coefficient of determination ($R^2$), Pearson correlation (Pearson *r*), and Spearman correlation (Spearman *r*) assess regression accuracy—together providing a comprehensive view of model effectiveness in the chart. Confidence intervals were estimated via nonparametric bootstrap resampling at the participant level (10,000 iterations; 95% confidence intervals) to quantify sampling variability given the moderate sample size.

In contrast, Non-Mirror evaluations based on life history interviews obtain lower predictive performances, consistent with the more challenging task of inferring symptoms from less structured narratives (Table 1b). GPT-4 Turbo reached an F1 score of 0.52 and Jaccard similarity of 0.34, and GPT-4o and LLaMA-3 70B also achieved similar levels. Yet despite lower values, Non-Mirror evaluations still captured substantial information about participants' overall depressive severity. For instance, GPT-4 Turbo achieved an $R^2$ of 0.27, Pearson correlation of 0.57 and Spearman correlation in 0.56 in the Non-Mirror condition—indicating strong alignment between predicted and actual total symptom counts, even when individual symptom predictions were imprecise. The moderate Pearson correlation (0.57) reflects a linear association between predicted and actual symptom totals, and the Spearman correlation (0.56) further suggests that the model preserves the rank ordering of severity across participants. This pattern held across models: GPT-4o and LLaMA 3-70B both achieved similar Pearson values (0.56 and 0.63, respectively), with slightly lower $R^2$ scores (0.16). Across models, Mirror conditions yielded narrower bootstrap confidence intervals. In contrast, Non-Mirror conditions exhibited wider confidence intervals for correlation-based metrics, indicating greater variability in the estimated associations.

| | Mirror - PHQ | | Non-Mirror-PHQ | |
|---|---|---|---|---|
| | Pearson | Spearman | Pearson | Spearman |
| GPT4 | 0.53 | 0.61 | 0.55 | 0.55 |
| GPT4o | 0.54 | 0.62 | 0.49 | 0.53 |
| LLaMA | 0.50 | 0.58 | 0.51 | 0.47 |

**Table 2.** Correlation between model-predicted severity and other clinical measures across condition. We report Pearson and Spearman correlation coefficients between model-predicted depression severity and self-reported PHQ-9 scores under both structured diagnostic (Mirror) and life history (Non-Mirror) interview conditions. All models demonstrate strong correlations.

When evaluated against an independent assessment, the apparent advantage of models in Mirror conditions diminished. Specifically, the model was additionally prompted with the original nine PHQ-9 scale items and their corresponding options. Table 2 reports correlations between model-predicted severity and PHQ-9 self-reported depression scores. For GPT-4 Turbo, the correlation was nearly identical across Mirror (Pearson r = 0.53, and Spearman r = 0.61) and Non-Mirror conditions (Pearson r = 0.55 and Spearman r = 0.55). GPT-4o and LLaMA-3 70B showed similar patterns, with Mirror condition correlations overlapping or falling below those of the Non-Mirror condition. Importantly, the PHQ-9 itself correlated strongly with DSM scores (Pearson r = 0.58, Spearman r = 0.63), confirming convergent validity of the independent PHQ-9 measure. The fact that models in Mirror conditions lost their advantage when compared against this external criterion indicates that their inflated in-sample performance is driven by criterion contamination, which is the structural overlap between predictors and targets.

| | *$R^2$ after filtering with confidence* | | *Pearson Score after filtering with confidence* | |
|---|---|---|---|---|
| | Mirror | Non-Mirror | Mirror | Non-Mirror |
| GPT-4 | 0.81 (+1.3%) | 0.35 (+29.6%) | 0.90 (+0%) | 0.64 (+12.3%) |
| GPT-4o | 0.78 (+1.3%) | 0.34 (+112.5%) | 0.90 (+0%) | 0.62 (+10.7%) |
| LLaMA-3 | 0.73 (+4.3%) | 0.21 (+31.3%) | 0.89 (+1.1%) | 0.65 (+3.2%) |

**Table 3.** Posterior filtering improves robustness of LLM predictions
Posterior filtering was applied by revising low-confidence "yes" predictions (confidence < 0.5) to "no." This unidirectional correction suppressed spurious affirmative responses and enhanced robustness, especially for Non-Mirror conditions. In GPT-4 Turbo, the Mirror evaluation improved slightly whereas the Non-Mirror evaluation improved substantially.

We next examined whether model-generated confidence ratings could be used as a post hoc filtering signal to reduce weakly supported positive predictions (as shown in Table 3). When designing the prompt, we explicitly instructed the LLM to output not only a categorical decision ("yes" or "no"), but also a confidence score indicating how strongly the transcript evidence supported that judgment. This design was motivated by the desire to obtain an interpretable, model-generated index of evidence strength (Tao et al., 2024; Tian et al., 2023). We treat these ratings as evidence-strength signals rather than as direct measures of the models' internal uncertainty. Specifically, when the model predicts a "yes" response with a confidence score below a defined threshold (0.5 in our study), the prediction was revised to "no." This unidirectional correction was motivated by two factors: (1) our prompt strictly constrained the model to predict "No" when no evidence was present; and (2) LLMs usually exhibit a tendency toward affirmative responses when given weakly grounded inputs (Yin et al., 2023). For GPT-4 Turbo, for the Mirror condition, performance slightly improved with $R^2$ increasing from 0.80 to 0.81, and the Pearson correlation remaining at 0.90; in contrast, the Non-Mirror condition benefitted substantially, with $R^2$ increasing from 0.27 to 0.35 (+29.6%) and the Pearson correlation from 0.57 to 0.64 (+12.3%). Consistent improvements were also observed across other LLMs. These findings suggest that leveraging confidence-based filtering can effectively enhance prediction robustness—particularly in less-structured language contexts with higher reasoning pressure.

These findings reveal a consistent picture. Mirror conditions achieve near-perfect results when evaluated against DSM-5 scores derived from the same structured diagnostic interview framework, a pattern consistent with criterion contamination. Non-Mirror conditions, while showing lower item-level accuracy, still achieve total-level effect sizes that are meaningful in psychological evaluations, including against the independent PHQ-9 measurement. Confidence-based filtering improved both conditions, with substantially larger gains in the Non-Mirror condition.

To evaluate the external validity of the Mirror and Non-Mirror derived depression scores, we examined their associations with a broad set of established psychological measures spanning mental health symptoms, personality traits, substance use, and protective factors. Pearson product–moment correlations were computed between each external variable and the depression scores obtained from both Mirror and Non-Mirror assessment procedures.

(a)

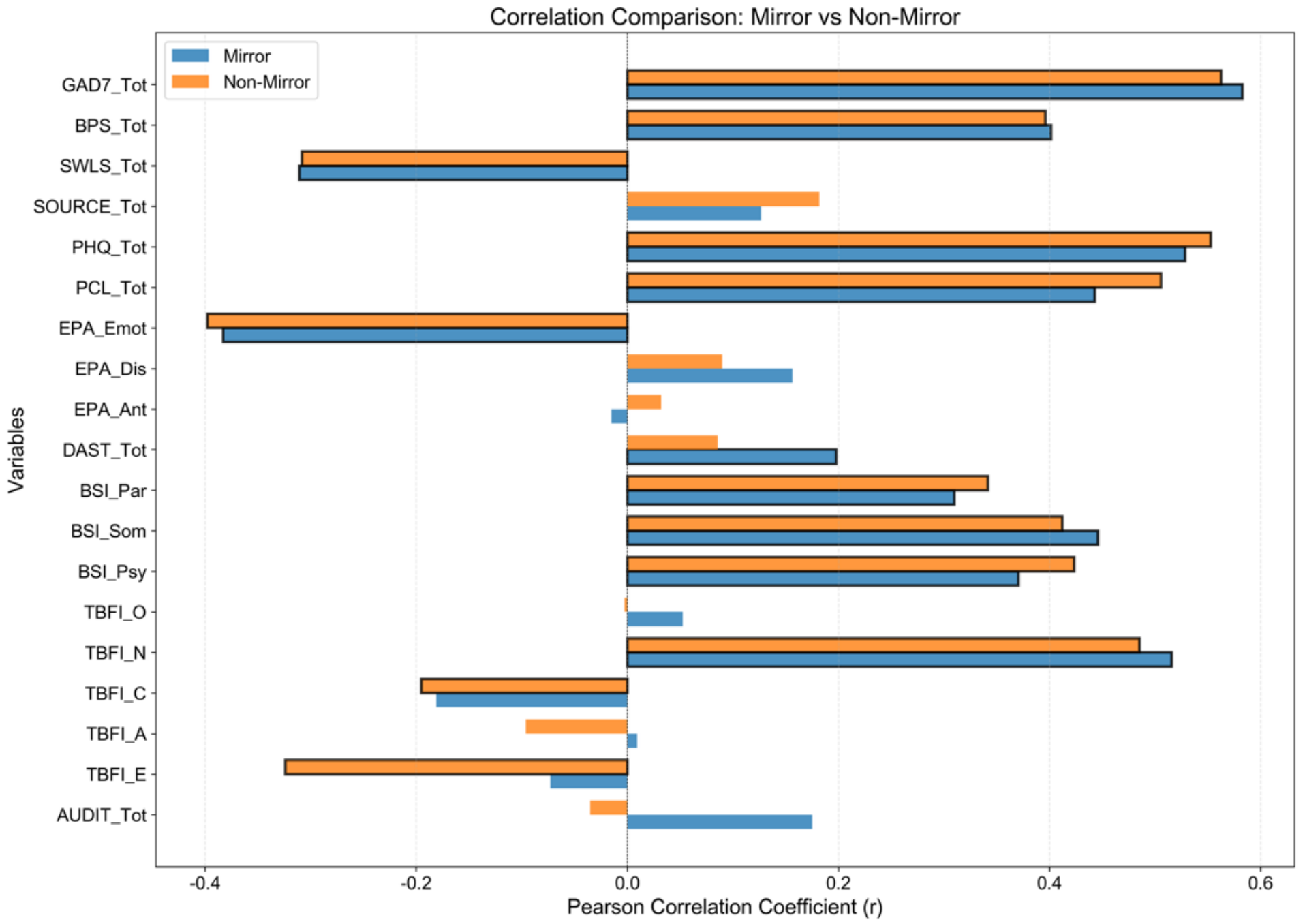

(b)

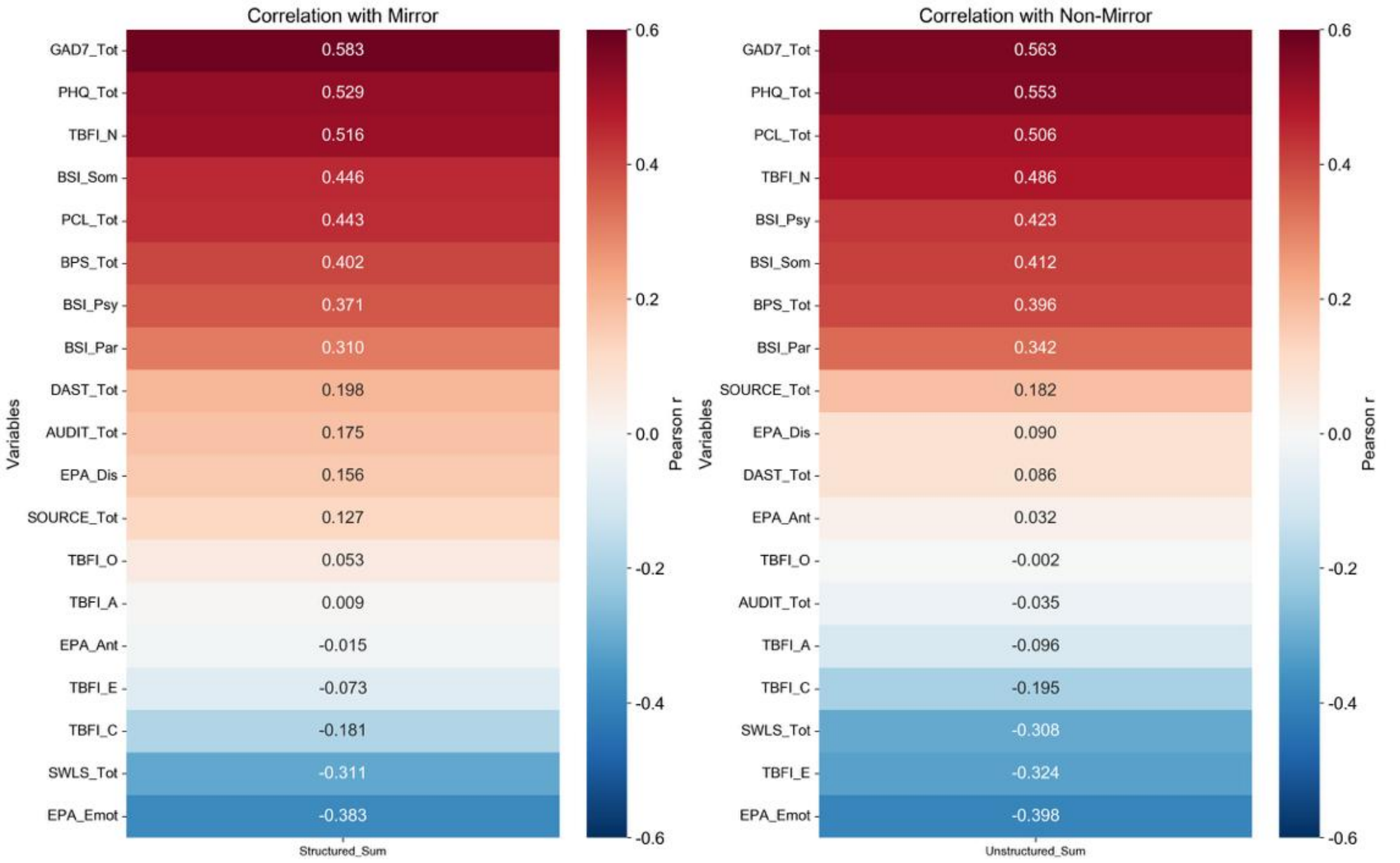

**Figure 3. Correlations between Mirror and Non-Mirror derived depression scores and external psychological measures.** GAD-7 = Generalized Anxiety Disorder-7; PHQ = Patient Health Questionnaire; PCL = PTSD Checklist for DSM-5; TBFI = BFI-2 = Big Five Inventory-2 (N = Neuroticism, O = Openness, A = Agreeableness, E = Extraversion, C = Conscientiousness); BSI = Brief Symptom Inventory (Som = Somatization, Psy = Psychoticism, Par = Paranoid Ideation); BPS = Borderline Pattern Scale; DAST = Drug Abuse Screening Test; AUDIT = Alcohol Use Disorders Identification Test; EPA = Elemental Psychopathy Assessment Super Short Form (Dis = Disinhibition, Ant = Antagonism, Emot = Emotional Stability); SOURCE = 4 item social dysfunction scale; SWLS = Satisfaction With Life Scale; Tot = total score. **Figure a.** presents a comparison of Pearson correlation coefficients between 19 external psychological

variables and depression scores derived from Mirror assessment and Non-Mirror assessment. For each variable, horizontal bars indicate the strength and direction of the correlation with each assessment method. Bars outlined in black denote statistically significant correlations ($p < .05$). **Figure b.** displays the same correlation coefficients using dual-panel heatmaps for Mirror (left panel) and Non-Mirror condition(right panel). Cell colors represent the magnitude and direction of correlations (Pearson's *r*), with warmer colors indicating positive correlations and cooler colors indicating negative correlations. Variables are sorted within each panel by absolute correlation strength to facilitate comparison of dominant association patterns across assessment methods. The specific definitions and statistical information regarding external psychological measures are detailed in the Appendix Table S3.

As shown in Figure 3, both assessment methods demonstrated similar depression construct and discriminant validity with core psychopathology measures. Depression (PHQ-9), anxiety (GAD-7), PTSD symptoms (PCL), and general symptom distress (BSI subscales) showed consistent moderate-to-strong positive correlations across both Mirror and Non-Mirror conditions. In contrast, theoretically relevant protective constructs, including emotional stability (EPA) and life satisfaction (SWLS), exhibited negative correlations, providing evidence of discriminant validity. Notably, differences were also observed. Model predictions from Mirror conditions showed relatively stronger associations with substance use indicators (e.g., DAST), whereas model predictions from Non-Mirror conditions exhibited stronger associations with extraversion and conscientiousness. These patterns suggest that while both approaches capture core depressive symptom severity, they may differentially reflect broader psychological dimensions.

**Robustness of Model Performance Across Symptoms**

Because the life history interview included questions about sleep and appetite, which overlap with DSM-5 depression symptom domains, we conducted a sensitivity analysis to examine whether Non-Mirror evaluation performance was driven primarily by these more directly observable symptom areas. Although they are not complete substantiations of the DSM symptoms—as they are in the structured interview—we decided to investigate whether the performance of the Non-Mirror condition was solely due to these overlapping symptom inquiries. We therefore systematically excluded each of the ten DSM-5 depression symptoms (Q1-Q10), and evaluated the absolute performance changes across both Mirror and Non-Mirror conditions. The left panel of our analysis (Figure 4) demonstrates the absolute Pearson correlation when each symptom is systematically removed from the full configuration. Both approaches maintain remarkable stability across all exclusion conditions. Our results show that removing Q3 (appetite/weight change) led to a very slight performance improvement of approximately 0.002, while excluding Q4 (sleep disorders) led to a very slight decline of approximately 0.04.

To quantify the relative importance of each symptom, we calculated importance scores using the formula: Importance Score = $(|\Delta Pearson_Structured| + |\Delta Pearson_LifeHistory|) / 2$, where delta values represent the change in Pearson correlation when excluding each symptom from the full configuration. This metric captures the average absolute change across both modeling paradigms, providing a unified measure of symptom influence. Most significantly, no single symptom emerged as disproportionately influential over the average of Mirror and Non-Mirrored modeling, with importance scores varying tightly between 0.003 (Q8) and 0.029 (Q1). The top-ranked symptoms – (Q1 (depressed mood): 0.029, Q6 (fatigue/energy loss): 0.023, Q9 (suicidal ideation): 0.023, Q4 (sleep disturbance):0.022) – were associated with relatively higher average performance impact when excluded. This indicates that both Mirror and Non-Mirror evaluations rely on the broad linguistic signals rather than a small set of overlapping specific symptoms. In the case of the two depression symptoms that were queried in the life history interview (sleep and appetite), these results indicate that these symptoms did not drive the predictive validity of the Non-Mirror score more than the others.

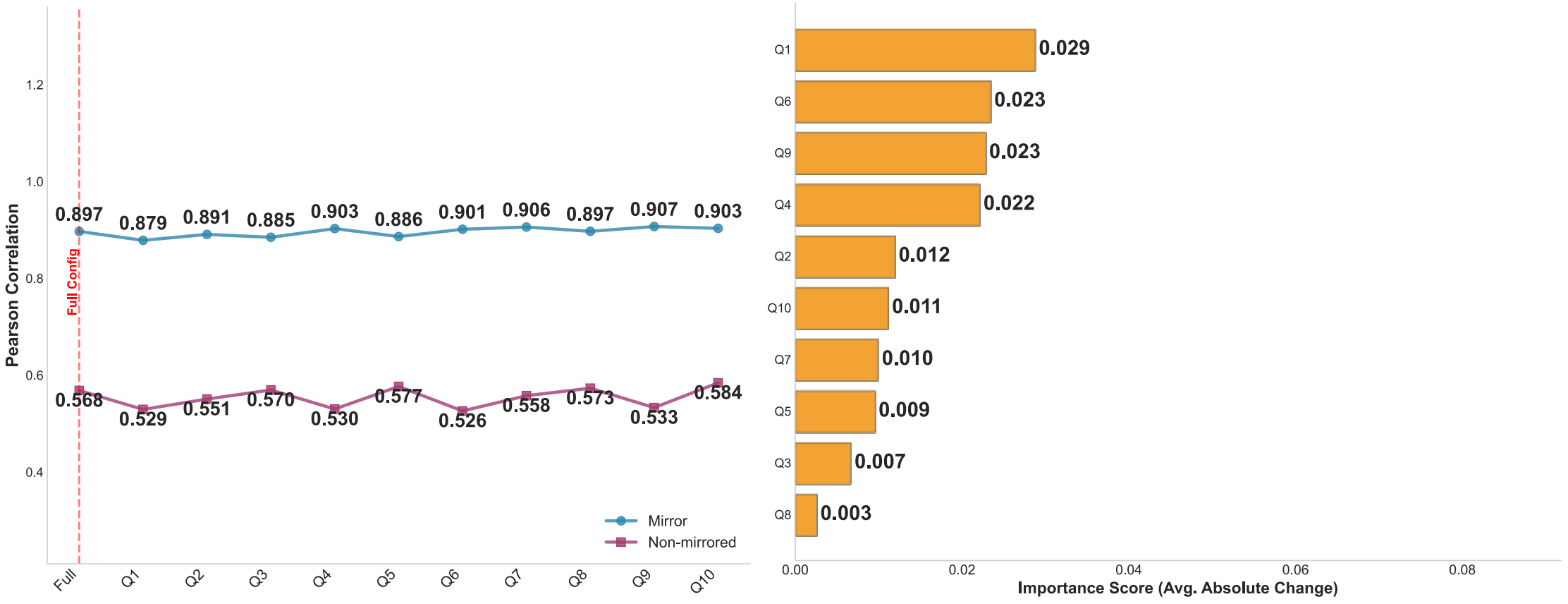

**Figure 4. Leave-One-Out Analysis of Depression Symptoms.** This analysis evaluates the contribution of each depression symptom (Q1-Q10) by excluding one symptom at a time and measuring the resulting performance changes. **(Left)** Absolute Pearson correlation trajectories show that both structured (blue circles) and unstructured (purple squares). **(Right)** Symptom importance ranking based on average absolute change.

## Topic Structures Differentiate Mirrored and Non-Mirror Conditions

We next used topic modeling as an interpretability and error-analysis tool to examine the transcript evidence underlying positive model predictions. The goal of this analysis was to characterize how symptom-related language differed across the Mirror and Non-Mirror conditions, and to identify types of life history language that may contribute to accurate versus inaccurate Non-Mirror predictions. Thus, the topic analysis was not intended as an additional performance metric, but as a way to understand what kinds of semantic evidence the models used when inferring depressive symptoms from different language contexts. We extracted 364 model-selected evidence utterances for positive DSM-based depressive symptom predictions from the Mirror condition and 573 from the Non-Mirror condition. Within the Non-Mirror subset, we further identified 249 *true* positive and 331 *false* positive utterances based on ground-truth annotations. Topics in the Mirror interview subset (Figure 5a) closely mirrored the DSM-5 symptom dimensions. The heatmap shows semantically coherent themes centered on DSM-5 depression symptoms such as sleep, appetite, suicide, cognitive difficulties, and guilt. The corresponding clusters in the UMAP graph were compact and well-separated, suggesting that the structured prompts elicited symptom-specific language. In contrast, the Non-Mirror interview subset exhibited broader and more overlapping topics spanning both symptom-related (e.g., sleep, appetite) and more contextual content including academic pressure, evaluation of personal change, and uncertainty about the future (Figure 5b). Coherent symptom-relevant or context-relevant topics from the Non-Mirror condition suggest meaningful psychological signal. The UMAP projection illustrates that utterances in life history interviews form less tightly bound and more diffuse clusters, reflecting the greater linguistic variability inherent in free-form dialogue. However, this variability does not preclude interpretability: the model was still able to extract symptom-relevant content, even when participants used more naturalistic or indirect expressions.

a

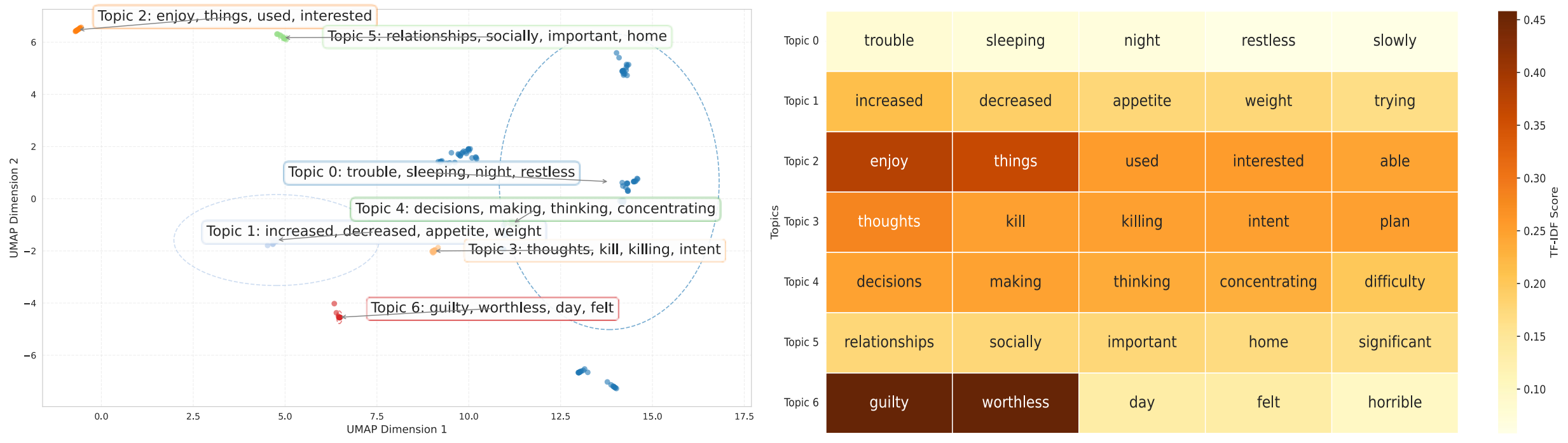

**Figure 5a left.** Mirror condition -- Semantic clustering of symptom-Positive utterances in structured diagnostic interviews. **Figure 5a right.** Keyword heatmap of topic clusters in structured diagnostic interviews.

b

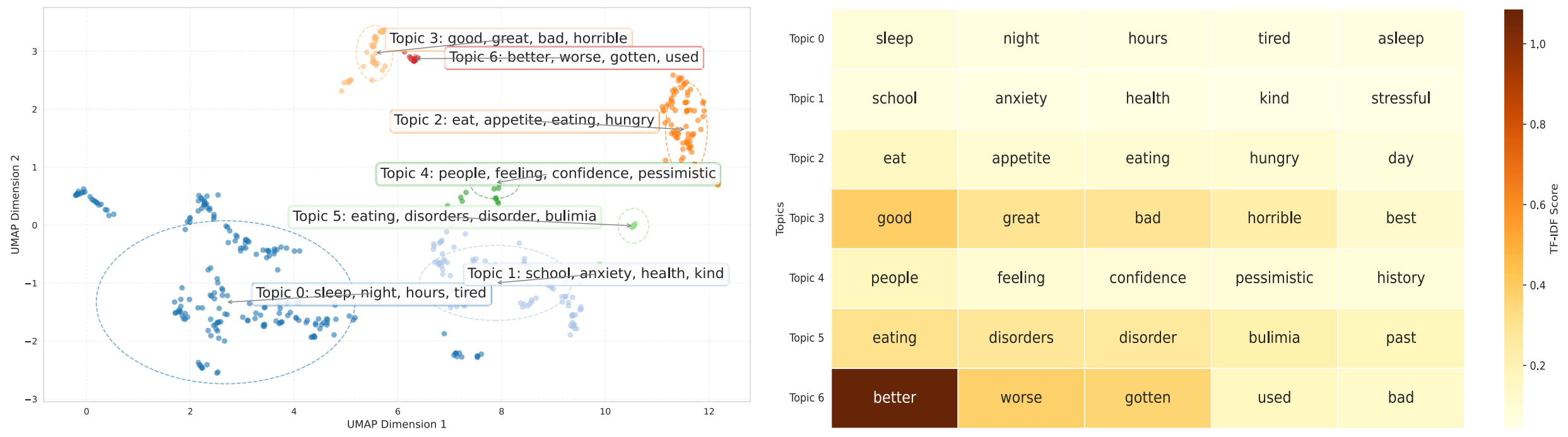

**Figure 5b left.** Non-Mirror condition -- Semantic clustering of symptom-Positive utterances in life history interviews. **Figure 5b right.** Keyword heatmap of topic clusters in life history interviews.

*Note.* The left panels display TF-IDF-based heatmaps of the top 5 keywords for each discovered topic, extracted from the BERTopic model and ordered by topic index. Color intensity reflects each keyword's relative TF-IDF weight, indicating its salience within the topic. The right panels present two-dimensional UMAP projections of sentence embeddings, where each point represents a model-selected utterance and is colored by its assigned topic. Topic labels are consistent with those from the original 5D topic space, enabling interpretable 2D visualization without altering the underlying topic assignments.

**Figure 5. Comparing Depression-Positive Topic Structures Across Mirror and Non-Mirror Conditions.**

We present qualitative examples from the Non-Mirror evaluation, highlighting linguistic features that led to different outcomes (Table 4). In the life history interview, LLMs can infer symptom presence or absence indirectly from relevant questions and responses not designed to probe symptoms. In a true negative for anhedonia, the model correctly inferred the absence of diminished interest from a casual question about leisure ("What do you like to do for fun?"), based on expressions of ongoing enjoyment and social motivation. In a true positive, it accurately detected fatigue from explicit mentions of chronic fatigue and managing energy—even though the question concerned life challenges, not symptoms. In a false positive, it over-interpreted normal behavioral variability, reading fluctuating appetite as symptom evidence. The false negative is instructive: the respondent described eating healthily and not being a picky eater—language that carries no appetite-disturbance signal—so the model's negative prediction was a correct reading of the available input. The discrepancy arose because the same respondent endorsed appetite change in the past two weeks in the structured interview. Such cases show that Non-Mirror predictions are limited when symptoms are absent, indirect, or temporally ambiguous in the narrative.

To further characterize how such symptom-relevant content is identified within Non-Mirror interviews, we conducted a symptom-level analysis examining the specific interview contextual cues supporting DSM-5 criteria. As summarized in Table S2 in the Appendix, Non-Mirror evaluations grounded their predictions in related inference sources when the symptom was not directly referred to, such as academic functioning, life stressors, and prior mental health experiences. Notably, high average confidence was observed across symptom domains, including sensitive constructs such as suicidality.

| Prediction Categories | DSM-5 Criterion & Reasoning (Rephased Summary) | Most Relevant Questions & Key Evidence (Partial) |
|---|---|---|
| **True Negative** | **Loss of interest or pleasure in most activities (Anhedonia):** The interviewee expressed clear and sustained interest in activities such as club soccer, skiing, and academic work. The LLM highlighted enjoyment in both physical and intellectual engagement, noting no sign of reduced pleasure or detachment. | **Key Evidence:**<br>“I'm in club soccer.”<br>“I just signed up for a skiing trip.”<br>“I love working with people…”<br>“Pre-med stuff that kind of excites you.”<br>**Most Relevant Question:**<br>“So what do you like to do for fun?” |
| **False Negative** | **Appetite or weight change:** The interviewee describes a stable appetite and healthy eating habits, with no mention of unintentional weight gain or loss. Their responses suggest no disordered eating or distress related to appetite. | **Key Evidence:**<br>“I'm not a picky eater at all. So I'll eat whatever someone gives me.”<br>“I tend to lean more on the healthier side.”<br>“I used to work out every single day.”<br>**Most Relevant Question:**<br>How would you describe your appetite? |
| **True Positive** | **Fatigue / Loss of energy:** The interviewee explicitly mentions chronic fatigue and the need to manage limited energy. This directly aligns with DSM-5 criteria regarding persistent tiredness. | **Key Evidence:**<br>“I do have a lot of chronic fatigue… I have to make sure I’m reserving energy so I can get to class.”<br>**Most Relevant Question:**<br>“What are the three biggest challenges in managing your life right now?” |
| **False Positive** | **Appetite or weight change:** The model identified fluctuating appetite patterns as indicative of | **Key Evidence:**<br>“There are days where I eat a lot.”<br>“Some days I just drink coffee.” |

| | |
|---|---|
| the symptom. The interviewee described alternating between overeating and minimal intake, such as only drinking coffee. This variability was interpreted as a signal of appetite disturbance. | **Most Relevant Question:** "And how would you describe your appetite?" |

**Table 4.** Examples of four prediction categories from the Non-Mirror condition. This table illustrates four examples from Non-Mirror condition, each corresponding to a different prediction category (True Positive, True Negative, False Positive, False Negative) for individual DSM-5 symptom criteria.

**Non-Mirror Condition Topics by True and False Positive**

Among positive Non-Mirror predictions, we further compared utterances associated with true-positive and false-positive classifications. This analysis was intended to clarify which kinds of life history language supported valid symptom inference and which kinds of language were more likely to produce erroneous positive predictions. In the true-positive subset (Figure 5a), topic modeling reveals semantically coherent clusters centered on broad clinically meaningful and sample specific content, as well as symptom specific topics. Themes reflected topics related to medication (e.g., medication, ADHD, regimen), school (e.g., semester, school, work, teach), and qualitative descriptions (e.g., good, bad, better, scary) as well as recognizable symptom domains such as sleep problems, appetite changes, and depressed mood. UMAP projections showed well-separated clusters, suggesting that when the Non-Mirror condition makes accurate predictions in life history interviews, it is based on relevant areas related to treatment, life circumstance descriptions, and sample-specific considerations (e.g., descriptions of school and work for this sample) as well as specific diagnostic symptom expressions.

a

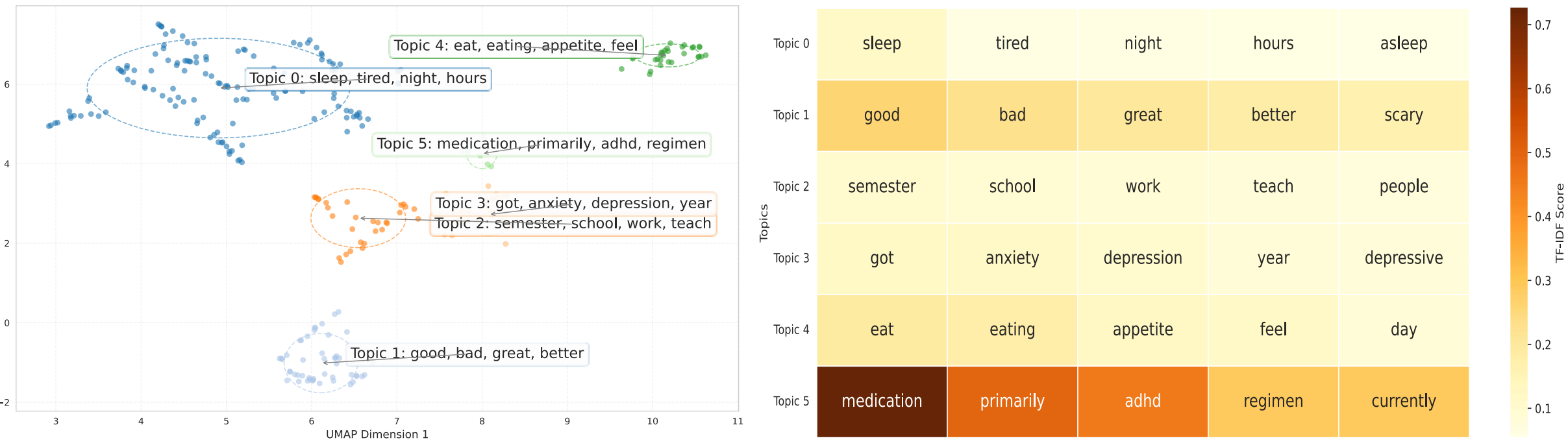

**Figure 6a left.** Non-Mirror -- Semantic clustering of symptom-True-Positive utterances in life history interviews. **Figure 6a right.** Keyword heatmap of topic clusters in life history interviews.

b

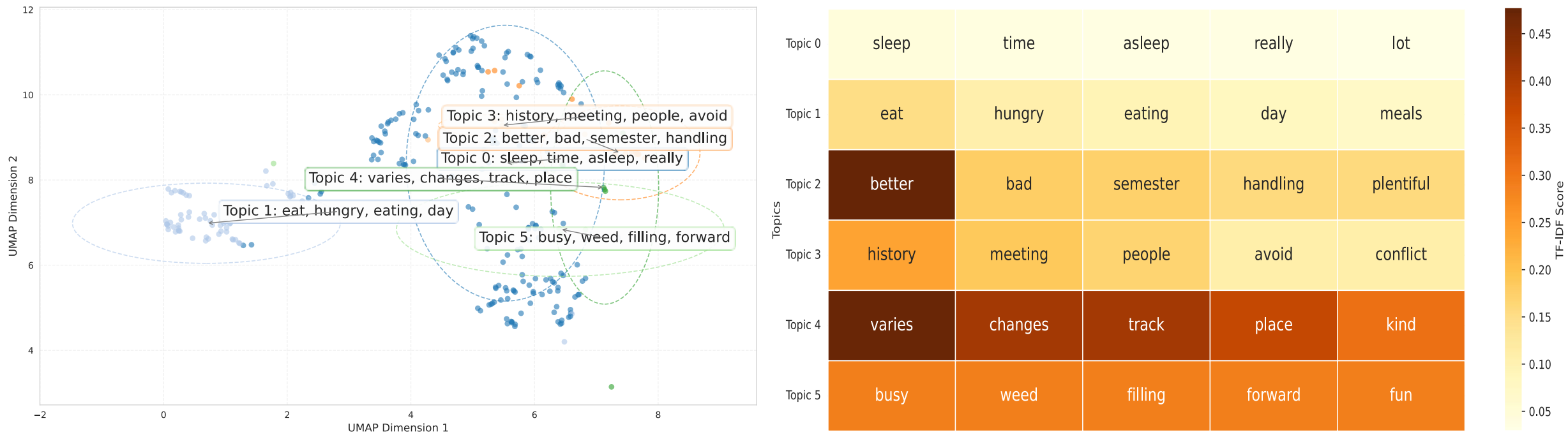

**Figure 6b left.** Non-Mirror -- Semantic clustering of symptom-False-Positive utterances in life history interviews. **Figure 6b right.** Keyword heatmap of topic clusters in life history interviews. *Note.* The left panels display TF-IDF-based heatmaps of the top 5 keywords for each discovered topic, extracted from the BERTopic model and ordered by topic index. Color intensity reflects each keyword's relative TF-IDF weight, indicating its salience within the topic. The right panels present two-dimensional UMAP projections of sentence embeddings, where each point represents a model-selected utterance and is colored by its assigned topic. Topic labels are consistent with those from the original 5D topic space, enabling interpretable 2D visualization without altering the underlying topic assignments.

**Figure 6. Comparing True and False Positive Topic Structures For Depression Predictions in the Non-Mirror Condition**

In contrast, the semantic structure of false-positive utterances (Figure 6b) was more diffuse, although the topics were coherent and interpretable. Themes in this group described getting better (topic 2), meetings (topic 3), and change (topic 4). The UMAP plots showed overlaps across thematic boundaries, indicating less pronounced semantic distinctions. Several of these themes point to a specific failure mode: topics such as getting better and change describe past or improving states, suggesting that the model sometimes read resolved or historical distress as currently present. This pattern identifies temporal anchoring as an important target for future Non-Mirror condition refinement.

To probe differences between true and false positives within the Non-Mirror condition further, we present representative keywords uniquely associated with true positive and false positive predictions, along with representative utterances (Table 5). In true-positive predictions, keywords such as scary and hard are consistently associated with emotional, symptom-expressive language, symptom-expressive language. Utterances containing these terms often described overwhelming thoughts, difficulty functioning, or explicit distress—e.g., “I was super depressed…” or “it's hard to calm it down.” These reflect clinically aligned semantics that anchor the model’s accurate prediction. In contrast, false-positive predictions were characterized by terms such as “history”, and “need.” For example, phrases such as “I need more” or “I can be a very difficult person to get along with” may convey difficulties that do not meet DSM-5 depression symptom thresholds, but lead the model to predict depressive symptom presence. The former aligns with the findings from Table 4 showing that discussion of a history of depressive symptoms may “trick” the model to predict the presence of current symptoms. Future development should integrate time-sensitive reasoning modules that help LLMs correctly anchor symptoms to a relevant time window. The latter suggests that discussion of need may also trick the model to predict current depressive symptoms. All of this is useful information to further improve Non-Mirror condition prompts in the future.

| Prediction Type | Representative Keyword | Representative Utterances |
|---|---|---|
| True Positive | scary | 1. “I'm finally realizing that I'm not a child anymore. and that's scary.”<br>2. “especially since I'm [redacted] turning 20… that's scary. It's |
| | | so scary… I was super depressed to the point where I was having suicidal thoughts.” |
| True Positive | hard | 1. “I always feel like I am on alert… It's a bit hard for me to turn that off.”<br>2. “It's just hard to calm it down… so many to-do lists going through my brain.” |
| False Positive | history | 1. “we've noticed a history of mental illness in my family.”<br>2. “I am a very, I can be a very difficult person to get along with just because of my history.” |
| False Positive | need | 1. “I need more. I think I need more.”<br>2. “just needed tutoring and more help.”<br>3. “I like going to counseling services when I know I need it.” |

**Table 5.** Examples for some representative keywords and related representative utterances in the Non-Mirror condition. Representative keywords are the keywords uniquely associated with True Positive and False Positive predictions.

Overall, these findings illustrate that Non-Mirror evaluations, even when producing imperfect symptom-level classifications, engage with language features that are psychologically meaningful and clinically relevant. Whereas Mirror evaluations achieve artificially high performance through criterion contamination, Non-Mirror evaluations extract information from naturalistic dialogue that reflects both symptom expressions and the broader context. This broader interpretive reach, consistent with the more diverse thematic structures identified in topic modeling, suggests that Non-Mirror evaluations offer both practical utility for prediction and potential scalable deployment in real-world mental health assessment. With further validation across domains, Non-Mirror evaluations may have particular utility in areas such as routine clinical settings, early detection, and broad population screening contexts.

## Discussion

Recent advances in large language models (LLMs) have created new opportunities for language-based assessment of depressive symptoms. However, the validity of these approaches depends not only on model performance, but also on the relation between the language sample used as input and the clinical criterion being predicted. In this study, we compared Mirror and Non-Mirror evaluation approaches to examine how assessment-elicited language and broader life history language differ as inputs for depression prediction. This distinction is important because strong model performance in psychological assessment can be inflated when the input language already contains information elicited by the same prompts, questions, or diagnostic criteria used to define the target label.

As found previously, Mirror evaluations achieved near-perfect effect sizes when predicting depression from structured diagnostic interviews that explicitly probed DSM-5 criteria. These findings further demonstrate that LLMs can reliably identify symptom expressions when such language is directly elicited. However, our results also show that this performance is inflated by criterion contamination: the input language in these interviews closely mirrors the outcome labels. When this contamination advantage was removed, for example by comparing model predictions against independent self-reported PHQ-9 scores, the apparent superiority of Mirror evaluations diminished substantially. Mirror and Non-Mirror evaluations showed similar correlations with PHQ-9 scores, suggesting that the higher $R^2$ of Mirror approaches was partly attributable to overlap with the diagnostic structure rather than broader generalizability.

Mirror evaluations should be conceptualized as *reliability* tests rather than validity tests. In the Mirror evaluation condition, the LLM is acting as a second scorer of the structured clinical interview. Coefficients derived from Mirror conditions ($r$ = .70-.90) are firmly in the range of reliability coefficients. Monomethod convergent validity coefficients are usually $r$ > .50 and cross-method validity at around $r$ > .30 (Clark & Watson, 2019; Funder & Ozer, 2019; Meyer et al., 2001). Mirror coefficients thus sit in the reliability band, well above the range that genuine validity estimates occupy—reinforcing that they index agreement between scorers of one procedure rather than convergence across methods. Viewed against these benchmarks, the Non-Mirror validity coefficients we observed (Pearson *r*s = .56-.63 across the three models) are striking: they are genuinely cross-method estimates, yet they exceed not only the cross-method threshold (~.30) but the monomethod threshold (~.50) as well. Conceptualizing the difference between Mirror (reliability) versus Non-Mirror (validity) evaluation conditions could be helpful in the developing field of psychometric evaluation of LLMs for depression assessment.

Further, although Mirror evaluations may be useful in narrowly defined contexts such as brief screenings for major depression, their broader application is limited. Real clinical encounters rarely involve administering all diagnostic questions verbatim across multiple disorders; instead, clinicians typically rely on open-ended discussions. In general practice, where multiple conditions must be assessed, structured Mirror interviews would be inefficient, often taking longer than standard self-report instruments. Thus, expanding Mirror evaluations undermines a central promise of AI in healthcare—improving efficiency and scalability.

In contrast, Non-Mirror evaluations showed lower symptom-level classification performance, as expected given the more difficult task of inferring depressive symptoms from less structured life history language. However, their subject-level associations with DSM-5 symptom counts and PHQ-9 scores remained meaningfully large by psychological assessment standards (Funder & Ozer, 2019; Meyer et al., 2001). They evaluated text from naturalistic, semi-structured life history interviews, which more closely resembles actual clinical encounters. Non-Mirror evaluations captured clinically meaningful signals not only from occasional explicit references to depressive symptoms but also from broader narratives involving life events, situational stressors, and psychosocial context. Topic modeling revealed that Non-Mirror predictions were informed by themes such as academic pressure, social comparison, and uncertainty about the future—factors that are not explicitly diagnostic but nevertheless shape daily functioning and vulnerability to depression. This ability to recognize distress in a wider range of contexts suggests that LLMs used in Non-Mirror conditions may enhance both clinical utility and scientific understanding.

Despite these strengths, Non-Mirror evaluations also revealed a characteristic limitation. Error analyses showed that false positives frequently arose from references to past or contextual improving states—language about distress that had since resolved—which the model interpreted as currently present.  This pattern points to temporal anchoring as a specific weakness: the models detected symptom-relevant content but failed to localize it in time. Improving temporal and contextual reasoning is therefore a clear priority for developing Non-Mirror approaches. A second source of apparent error is structural rather than inferential: because life-history interviews were not designed to systematically probe every depressive symptom, a symptom recorded in the structured interview was sometimes never addressed in the narrative. In these cases the model's negative prediction reflects an accurate reading of the available language rather than an inference failure, yet it is scored as an error against the criterion (e.g., the appetite-change example in Table 4). An unknown portion of Non-Mirror item-level misclassifications may therefore stem from uneven symptom coverage in the narrative rather than model error. This consideration applies to item-level classification metrics rather than to the subject-level validity coefficients, but it suggests that item-level accuracy may understate the model's ability to detect symptom that are actually expressed in the language.

Overall, our findings suggest that while Mirror evaluations demonstrate the feasibility of language-based AI for detecting explicitly elicited symptoms, their high performance reflects criterion contamination rather than generalizable validity. Non-Mirror evaluations, although less precise on item-level metrics, align more closely with real-world clinical language and capture a broader range of meaningful signal. Notably, their subject-level coefficients ($r$ = .56-.63) reach levels conventionally considered strong even for monomethod convergent validity ($r \approx .50$)—despite being cross-method estimates, which are typically expected to be lower still. Future progress will depend on improving the temporal and contextual reasoning of Non-Mirror approaches, so they can extract more reliable and clinically interpretable information from naturalistic discourse. By doing so, language-based AI could realize its dual potential: improving the efficiency of psychological assessment and deepening our understanding of the complex ways in which depression is expressed in human language.

**Limitations**

Several considerations are relevant when interpreting these findings. First, larger and more diverse samples will be necessary to more precisely characterize the stability and generalizability of these effects, including whether the Mirror/Non-Mirror pattern holds consistently across demographic groups and levels of clinical severity. Second, additional interview formats will be needed to identify where Non-Mirror LLM assessment is most useful—for example, in primary care visits, routine psychotherapy, or crisis line calls. Third, this study evaluated Mirror versus Non-Mirror approaches for a single construct (depression); whether criterion contamination inflates performance similarly for other clinical constructs (e.g., anxiety, trauma) remains to be tested. Finally, the present study evaluated these approaches using only prompting of LLMs. Replication and extension using other methods of working with LLMs (Brickman et al., 2025), such as fine-tuning, will be important for understanding how the distinction operates across modeling approaches that may carry different future clinical implications.

**Conclusions**

Language-based AI has substantial potential for depression assessment, but its validity depends on how the language sample relates to the criterion being predicted. This distinction is especially important when interpreting high performance. Mirror evaluations—in which the input language is elicited by the same diagnostic framework used to define the outcome—can achieve near-perfect performance, but such coefficients fall in the range of reliability rather than validity and may reflect criterion contamination rather than generalizable depression detection. Non-Mirror evaluations showed lower symptom-level accuracy, yet they addressed a more difficult and clinically relevant task: inferring depressive symptoms from broader life-history language not designed to reproduce DSM-5 criteria. Their cross-method coefficients nonetheless reached levels considered strong even for monomethod convergent validity. Distinguishing Mirror from Non-Mirror evaluations can help researchers and clinicians interpret language-model performance more appropriately, and the distinction applies wherever a criterion is derived from the same language used to predict it—a consideration extending beyond depression to language-based assessment broadly.

# Appendix

**Algorithm 1** Prompt Construction for DSM-5 Depression Symptom Inference

**Require:** Transcript text $T$, Interview type $I \in \{\text{structured}, \text{unstructured}\}$

**Ensure:** Prompt $P$ containing step-wise instructions and questions for LLM input

1: Define symptom questions $Q = [Q_1, Q_2, \ldots, Q_{10}]$, where:

- $Q_1$: "Has the individual felt down, sad, empty, or hopeless for most of the day, nearly every day?"
- $Q_2$: "Have they found themselves much less interested in most activities or unable to enjoy things they used to like, for most of the time?"
- $Q_3$: "Have they noticed a significant change (increase or decrease) in appetite almost every day or experience significant increase or decrease in weight without trying?"
- $Q_4$: "Have they had trouble sleeping almost every night (i.e., problems falling asleep, waking up too much, waking too early, or sleeping too much)?"
- $Q_5$: "Have they noticed talking or moving more slowly than is usual or being more restless or fidgety than usual nearly every day?"
- $Q_6$: "Have they felt tired or lacking energy nearly every day?"
- $Q_7$: "Have they more often felt worthless or excessively guilty?"
- $Q_8$: "Have they had trouble focusing, thinking clearly, or making decisions nearly every day?"
- $Q_9$: "Have they had frequent thoughts about death (not including fear of dying), or thought about/planned/tried to kill themselves?"
- $Q_{10}$: "Have these symptoms caused major distress or important difficulties in their daily life including occupationally, educationally, relationally, or socially, compared to before?"

2: **if** $I$ = structured **then**

3: Set *interview_context* to: "This is a diarized transcript of a mental health interview where a structured interview scale of ten items is administered to assess for DSM-5 major depressive disorder. Each question is asked first about its presence in the past and then about its presence in the past two weeks."

4: **else**

5: Set *interview_context* to: "This is a diarized transcript of a mental health interview where a clinical interview is being administered to assess mental health-relevant history. In the text, the interviewer asks the interviewee about background, relationships, health, and strengths."

6: **end if**

7: Construct prompt $P$ as the concatenation of:

- **Context:** "You are a clinical psychologist, analyzing interview transcripts to detect symptoms of depression according to DSM-5 criteria. Your task is to classify whether the provided transcript contains clear evidence of the symptom in the last two weeks mentioned in the given question."
- **Interview Context:** Insert *interview_context*
- **Task Instructions:**
  1. Identify Evidence: Determine whether there is direct or indirect evidence of the symptom in the last two weeks.
  2. Classify Confidence (1–9):
     - 7–9: Direct, clear statement confirming the symptom
     - 4–6: Symptom is mentioned but inferred from context
     - 1–3: Weak indication, requiring speculative inference
     - 9: If symptom is not mentioned, return `"Prediction": "No"` with `"Confidence": 9`
  3. Extract Evidence: Identify and list key utterances that support your reasoning.
  4. Contextualize: Determine the most relevant interviewer question that led to the answer.
- **Question:** Insert $Q_i$ from the list $Q$
- **Transcript:** Insert the full text $T$
- **Output Format:**

```
{
 "Prediction": "Yes" or "No",
 "Confidence": X,
 "Explanation": "...",
 "Key Utterances": ["..."],
 "Most Relevant Question": "..."
}
```

8: **return** Prompt $P$

**Figure S1. Prompt construction algorithm for DSM-5 depression symptom reasoning.** This Figure illustrates the full prompt and the construction pipeline for DSM-5 depression symptom

reasoning. The process involves inserting question-specific symptom prompts, providing clinical context, stepwise instruction for evidence identification and confidence classification, and specifying a structured output format to support LLM interpretability and hallucination reduction.

| Category | Value | Statistic / Percentage |
|---|---|---|
| Sample number | Sum | 110 |
| Age Distribution | Mean | 20.14 |
| | Standard Deviation | 2.43 |
| | Minimum | 18 |
| | Median | 20 |
| | Maximum | 36 |
| GPA Distribution | Mean | 3.32 |
| | Standard Deviation | 0.47 |
| | Minimum | 1.72 |
| | Median | 3.40 |
| | Maximum | 4 |
| Gender Identity Distribution | Woman | 66.36% |
| | Man | 30.91% |
| | Transgender Woman | 0.91% |
| | Transgender Man | 0.91% |
| | Non-Binary | 0.91% |
| Sex Distribution | Female | 68.18% |
| | Male | 31.82% |
| Sexual Orientation Distribution | Heterosexual/Straight | 82.73% |
| | Gay/Lesbian | 3.64% |
| | Bisexual | 9.09% |
| | Queer | 0.91% |
| | Asexual | 1.82% |
| | aromantic | 0.91% |
| | Prefer not to reply | 0.91% |
| Race Distribution | Single Selection | 95.45% |
| | Multiple Selection | 4.55% |
| | American Indian/Alaskan Native | 1.82% |
| | Asian | 5.45% |
| | Black/African American | 26.36% |
| | Native Hawaiian/Pacific Islander | 1.82% |
| | White/Caucasian | 67.27% |
| | Other or unknown | 2.73% |
| Ethnicity Distribution | Not of Hispanic/Latino/a/x Descent | 91.70% |
| | Hispanic/Latino/a/x Descent | 8.26% |
| Marital Status Distribution | Single | 96.36% |
| | Married | 1.82% |
| | Cohabitating | 1.82% |
| Family Income Distribution | Under $20,000 | 1.82% |
| | $20,000 - $39,999 | 3.64% |
| | $40,000 - $59,999 | 7.27% |
| | $60,000 – $79,999 | 22.73% |
| | $80,000 - $99,999 | 14.55% |
| | $100,000 - $119,999 | 13.64% |
| | $120,000 - $139,999 | 9.09% |
| | $140,000 or more | 27.27% |
| DSM-5 Labels Distribution | Minimum | 0.0 |
| | Maximum | 8.0 |
| | Mean | 1.83 |
| | Variance | 3.41 |

**Table S1.** Demographic and Socioeconomic Characteristics of Data (N = 110). This table summarizes the demographic, academic, and socioeconomic characteristics of the 110 subjects. It includes distributions of age, GPA, gender identity, sex, sexual orientation, race, ethnicity, marital status, and family income. Means, standard deviations, and ranges are reported for continuous variables (age and GPA), while categorical variables are presented as percentages of the total sample. *Note: Participants were allowed to select more than one racial category. As a result, percentages for specific racial groups may sum to more than 100%.*

| DSM-5 Criterion No. | DSM-5 Symptom Description | Sample Size | Directly Asked Rate | Primary Inference Source | Average Confidence |
|---|---|---|---|---|---|
| 1 | Depressed mood most of the day, nearly every day (e.g., feeling sad, empty, or hopeless) | 45 | 8.9% | "What do you like to do for fun?" /"How's school going for you currently?" | 8.53 |
| 2 | Markedly diminished interest or pleasure in all, or almost all, activities (anhedonia) | 106 | 2.8% | "What do you like to do for fun?" | 8.86 |
| 3 | Significant weight loss or gain, or decrease or increase in appetite | 104 | 97.1% | Mostly directly asked (e.g., "How about your appetite?") | 8.14 |
| 4 | Insomnia or hypersomnia nearly every day | 105 | 97.1% | Mostly directly asked (e.g., About how many hours a night do you get?) | 7.66 |
| 5 | talking or moving more slowly than is usual or being more restless or fidgety | 19 | 5.3% | “How about your sleep quality?” / “Do you find exercise important?” | 8.63 |
| 6 | Fatigue or loss of energy nearly every day | 40 | 0% | “How would you describe your sleep quality?” | 8.03 |
| 7 | Feelings of worthlessness or excessive or inappropriate guilt | 29 | 3.4% | “How has school been going for you?”/ “biggest challenges” | 8.31 |
| 8 | Diminished ability to think or concentrate, or indecisiveness | 44 | 2.3% | “What are your current school-related stressors?” / “any past or present educational problems” | 7.20 |
| 9 | Recurrent thoughts of death, recurrent suicidal ideation, or suicide attempt | 28 | 3.6% | “any significant life changes, stress”/ “ever seen a doctor or a mental health provider for an emotional or behavior disorder” | 8.68 |
| 10 | Symptoms cause clinically significant distress or impairment in social, occupational, or other important areas of functioning | 50 | 12.0% | “In what areas of your life are you having difficulty?”/ “biggest challenges” | 7.10 |

**Table S2.** Symptom-level inference patterns in non-mirror (life-history) interviews. For each DSM-5 major depressive disorder criterion, the table summarizes how symptom estimates were supported in the absence of direct questioning, including representative interview prompts serving as primary inference sources and the average confidence of model outputs. Despite low rates of explicit assessment for several symptoms—particularly sensitive domains such as suicidality—models identified relevant contextual cues and produced high-confidence estimates.

| Variable | N | Mean | Std | Min | Q1 | Median | Q3 | Max | Missing |
|---|---|---|---|---|---|---|---|---|---|
| **AUDIT_Tot** | 105 | 14.35 | 4.17 | 10.00 | 11.00 | 13.00 | 17.00 | 27.00 | 3 |
| **TBFI_E** | 108 | 16.61 | 4.07 | 7.00 | 14.00 | 16.50 | 20.00 | 24.00 | 0 |
| **TBFI_A** | 108 | 23.66 | 3.68 | 13.00 | 21.00 | 23.50 | 27.00 | 30.00 | 0 |
| **TBFI_C** | 108 | 21.97 | 4.71 | 8.00 | 19.00 | 22.00 | 25.00 | 30.00 | 0 |
| **TBFI_N**** | 108 | 17.64 | 5.20 | 6.00 | 14.00 | 17.50 | 22.00 | 30.00 | 0 |
| **TBFI_O**** | 108 | 22.97 | 4.28 | 14.00 | 19.00 | 23.00 | 27.00 | 30.00 | 0 |
| **BSI_Psy**** | 105 | 5.39 | 5.05 | 0.00 | 1.00 | 4.00 | 8.00 | 22.00 | 3 |
| **BSI_Som**** | 105 | 3.66 | 4.96 | 0.00 | 0.00 | 2.00 | 5.00 | 18.00 | 3 |
| **BSI_Par**** | 105 | 5.08 | 4.61 | 0.00 | 1.00 | 4.00 | 7.00 | 22.00 | 3 |
| **DAST_Tot**** | 105 | 14.56 | 4.34 | 10.00 | 11.00 | 14.00 | 16.00 | 32.00 | 3 |
| **EPA_Ant**** | 105 | 14.57 | 3.51 | 8.00 | 12.00 | 14.00 | 17.00 | 24.00 | 3 |
| **EPA_Dis**** | 105 | 11.59 | 4.71 | 5.00 | 7.00 | 11.00 | 15.00 | 24.00 | 3 |
| **EPA_Emot**** | 105 | 14.54 | 3.51 | 5.00 | 12.00 | 15.00 | 17.00 | 24.00 | 3 |
| **PCL_Tot**** | 105 | 18.26 | 14.73 | 0.00 | 6.00 | 15.00 | 28.00 | 67.00 | 3 |
| **PHQ_Tot**** | 108 | 5.56 | 4.82 | 0.00 | 2.00 | 4.00 | 8.00 | 23.00 | 0 |
| **SOURCE_Tot**** | 108 | 7.15 | 2.69 | 4.00 | 5.00 | 7.00 | 8.00 | 18.00 | 0 |
| **SWLS_Tot**** | 108 | 23.37 | 6.84 | 7.00 | 19.00 | 24.00 | 29.00 | 35.00 | 0 |
| **BPS_Tot**** | 105 | 23.30 | 8.02 | 12.00 | 17.00 | 23.00 | 28.00 | 56.00 | 3 |
| **GAD7_Tot**** | 105 | 7.82 | 5.56 | 1.00 | 4.00 | 6.00 | 11.00 | 23.00 | 3 |

**Table S3. Statistical Information of External Psychological Measures.** The 19 external variables represent established measures across psychopathology, personality, substance use, and well-being domains: **AUDIT_Tot** (alcohol use severity), **DAST_Tot** (drug use severity), **PHQ_Tot** (depressive symptoms) (Kroenke et al., 2001), **GAD7_Tot** (anxiety symptoms) (Spitzer el al., 2006), **PCL_Tot** (posttraumatic stress symptoms) (Blevins et al. 2015), **BSI_Psy** (psychoticism), **BSI_Som** (somatization), **BSI_Par** (paranoid ideation), **BPS_Tot** (borderline personality features) (Derogatis et al. 1983), **TBFI_N** (neuroticism), **TBFI_E** (extraversion), **TBFI_A** (agreeableness), **TBFI_C** (conscientiousness), **TBFI_O** (openness), **EPA_Ant** (antagonism), **EPA_Dis** (disinhibition), **EPA_Emot** (emotional stability) (Lynam et al. 2011), **SWLS_Tot** (life satisfaction), and **SOURCE_Tot** (source memory performance).